\documentclass{article}
\usepackage{amssymb}
\usepackage{xcolor}
\usepackage[implicit=false]{hyperref}

\long\def\comment#1{}

\newtheorem{assumption}{Assumption}
\newtheorem{theorem}{Theorem}

\newtheorem{corollary}[theorem]{Corollary}
\newtheorem{lemma}[theorem]{Lemma}

\newcommand{\Ass}[1]{Assumption~\ref{#1}}

\newcommand{\Eq}[1]{Equation~\ref{#1}}

\newcommand{\qed}{\mbox{$\Box$}}

\newcommand{\X}{X}
\newcommand{\x}{x}
\newcommand{\Db}{d}
\newcommand{\U}{{\bf X}}
\newcommand{\C}{{\bf x}}
\newcommand{\bY}{{\bf Y}}
\newcommand{\by}{{\bf y}}
\newcommand{\Pai}{{\bf Pa}_i}
\newcommand{\pai}{{\bf pa}_i}
\newcommand{\pail}{{\bf pa}_{il}}
\newcommand{\PaSi}{{\bf Pa}^m_i}
\newcommand{\paSi}{{\bf pa}^m_i}
\newcommand{\PaScii}{{\bf Pa}^{m_{c(i)}}_i}

\newcommand{\bX}{{\bf X}}
\newcommand{\bx}{{\bf x}}

\newcommand{\Bm}{m}
\newcommand{\Bmc}{m_c}
\newcommand{\Bmci}{m_{c(i)}}

\newcommand{\Bmone}{m_1}
\newcommand{\Bmtwo}{m_2}

\newcommand{\Bs}{s}
\newcommand{\Bsc}{s_c}
\newcommand{\Bsci}{s_{c(i)}}

\newcommand{\Bsone}{s_1}
\newcommand{\Bstwo}{s_2}

\newcommand{\hBs}{m^h}
\newcommand{\hBsc}{m^h_c}
\newcommand{\hBsci}{m^h_{c(i)}}

\newcommand{\hBsone}{m^h_1}
\newcommand{\hBstwo}{m^h_2}

\newcommand{\Fs}{{\cal F}_s}
\newcommand{\Fsone}{{\cal F}_{s1}}
\newcommand{\Fstwo}{{\cal F}_{s2}}

\newcommand{\Jarai}{J\'{a}rai }
\newcommand{\Jarais}{J\'{a}rai's }
\newcommand{\Aczel}{Acz\'{e}l }

\newcommand{\dXdY}[2]{\partial {#1} / \partial {#2}}

\newcommand{\Th}{\mbox{$\theta$}}

\newcommand{\Thi}{\Th_i}
\newcommand{\Thj}{\Th_j}
\newcommand{\TBs}{\Th_{m}}
\newcommand{\TBsc}{\Th_{mc}}
\newcommand{\TBsci}{\Th_{mc(i)}}
\newcommand{\TBsone}{\Th_{m1}}
\newcommand{\TBstwo}{\Th_{m2}}

\newcommand{\CTBs}{\Theta_{m}}
\newcommand{\CTBsone}{\Theta_{m1}}

\newcommand{\vecb}{{b}} 
\newcommand{\matW}{W} 
\newcommand{\matS}{S} 
\newcommand{\matM}{M} 
 
\newcommand{\Xvecbar}{\overline{{\bf x}}}
\newcommand{\Xvecsubl}{{\bf x}_l}
\newcommand{\vecmu}{\mu}
 
\newcommand{\tr}{\mbox{\it tr}} 
\newcommand{\amu}{\alpha_{\mu}} 
\newcommand{\aw}{\alpha_w} 
\newcommand{\awY}{\alpha_w -n + l}
 
\newcommand{\A}{W_{11}} 
\newcommand{\B}{W_{12}} 
\newcommand{\D}{W_{22}} 

\newcommand{\W}{\mbox{\cal Wishart}} 
\newcommand{\m}{N}
\newcommand{\vb}{|\!|} 

\title{Parameter Priors for Directed Acyclic Graphical 
Models and the Characterization of Several Probability 
Distributions}

\author{
Dan Geiger\\ 
geiger02@gmail.com\\
\\
David Heckerman\\
heckerma@hotmail.com}

\date{Annals October 2002 version with corrections and updates made June 2021}

\begin{document}

\maketitle

\begin{abstract}

\noindent We develop simple methods for constructing
parameter priors for model choice among
Directed Acyclic Graphical (DAG) models.  
In particular, we introduce several
assumptions that permit the construction of parameter
priors for a large number of DAG models from a small
set of assessments.  
We then present a method for directly computing the
marginal likelihood of every DAG model given
a random sample with no missing observations.
We apply this methodology to Gaussian DAG models 
which consist of a recursive set of linear regression models.  
We show that the only parameter prior
for complete Gaussian DAG models 
that satisfies our assumptions
is the normal-Wishart distribution.  
Our analysis is based on the
following new characterization of the Wishart distribution:
let $W$ be an $n \times n$, $n \ge 3$, positive-definite symmetric 
matrix of random variables
and $f(W)$ be a pdf of $W$.
Then, f$(W)$ is a Wishart distribution if and only if
$W_{11} - W_{12} W_{22}^{-1} W'_{12}$ is independent of 
$\{W_{12},W_{22}\}$
for every block partitioning $W_{11},W_{12}, W'_{12}, W_{22}$ of
$W$. 
Similar characterizations of the normal and normal-Wishart
distributions are provided as well.

\bigskip


\bigskip

\noindent
Corrections to the original text in \textcolor{red}{red} are
taken from J. Kuipers, G. Moffa, and D. Heckerman, Addendum on the
scoring of Gaussian directed acyclic graphical models. {\em Annals of
Statistics} 42, 1689-1691, Aug 2014. 
Other updates to the original are in \textcolor{blue}{blue}.

\end{abstract}

\section{Introduction} \label{sec:intro}

Directed Acyclic Graphical (DAG) models have increasing number of
applications in Statistics (Cowell, Dawid, Lauritzen, and
Spiegelhalter, 1999)\nocite{CDLS99} as well as in Decision Analysis
and Artificial Intelligence (Howard and Matheson, 1981; Heckerman,
Mamdani, and Wellman, 1995b; Pearl,
1988). \nocite{Howard81,Pearl88,HMW95cacm} A DAG model $\Bm = (\Bs,
\Fs)$ for a set of variables $\U=\{X_1,\ldots,X_n\}$ each associated
with a set of possible values $\mbox{D}_i$, respectively, is a set of
joint probability distributions for $\mbox{D}_{1} \times \cdots \times
\mbox{D}_{n}$ specified via two components: a structure $\Bs$ and a
set of local distribution families $\Fs$. The structure $\Bs$ for $\U$
is a directed graph with no directed cycles (i.e., a Directed Acyclic
Graph) having for every variable $X_i$ in $\U$ a node labeled $X_i$
with parents labeled by $\PaSi$.  The structure $\Bs$ represents the
set of conditional independence assertions, and only these conditional
independence assertions, which are implied by a factorization of a
joint distribution for $\U$ given by $p(\C) = \prod_{i=1}^n
p(\x_i|\paSi)$, where $\C=(x_1,\ldots,x_n)$ is a value for $\U$ (an
$n$-tuple) and $x_i$ is a value for $X_i$ and where $\paSi$ is the
value for $\PaSi$ as in $\C$.  When $\x_i$ has no incoming arcs in $m$
(no parents), $p(\x_i|\paSi)$ stands for $p(\x_i)$.  The local
distributions are the $n$ conditional and marginal probability
distributions that constitute the factorization of $p(\C)$. Each such
distribution belongs to the specified family of allowable probability
distributions $\Fs$.  A DAG model is often called a {\em Bayesian
network}, although the later name sometimes refers to a specific joint
probability distribution that factorizes according to a DAG, and not,
as we mean herein, a set of joint distributions each factorizing
according to the same DAG.  A DAG model is {\em complete} if it has no
missing arcs. Note that any two complete DAG models for $\U$ encode
the same assertions of conditional independence---namely, none.  Also
note that a complete DAG determines a unique ordering of the variables
in which $X_i$ precedes $X_j$ if and only if $X_i \rightarrow X_j$ is
an arc in this DAG.

In this paper, we assume that
each local distribution is selected from 
a family $\Fs$ which depends on a
finite set of parameters $\TBs \in \CTBs$
(a parametric family). The parameters for a local
distribution is a set of real numbers that 
completely determine the functional form of
$p(\x_i|\paSi)$ when $x_i$ has parents
and of $p(\x_i)$ when $x_i$ has no parents.
We denote by $\hBs$ the model hypothesis that the 
true joint probability distribution of $\U$ 
is perfectly represented by a structure $\Bs$ of a DAG model $\Bm$
with local distributions from $\Fs$---namely, 
that the joint probability distribution satisfies 
only the conditional independence assertions
implied by this factorization and none other.
Consequently, the true joint distribution for a DAG 
model $\Bm$ is given by,
\begin{equation} \label{eq:bn-def-p}
p(\C|\TBs,\hBs) = \prod_{i=1}^n p(\x_i|\paSi,\Thi,\hBs)
\end{equation}
where 
$\theta_1,\ldots \theta_n$ are subsets of $\TBs$.
Whereas in a general formulation of DAG models,
the subsets $\{\theta_i\}_{i=1}^n$ could possibly overlap 
allowing several local distributions to have common parameters,
in this paper, we shall shortly exclude this possibility
(Assumption~\ref{ass:pi}). Note that $\theta_m$ denotes the
union of $\theta_1, \ldots,\theta_n$ for a DAG model $m$.

We consider the Bayesian approach when
the parameters $\TBs$ and the
model hypothesis $\hBs$ are uncertain
but the parametric families are known.
Given data
$\Db=\{\C_1,\ldots,\C_{\m}\}$, a random sample from $p(\C|\TBs,\hBs)$
where $\TBs$ and $\hBs$ are the true parameters and model
hypothesis, respectively, we can compute the posterior probability of
a model hypothesis $\hBs$ using
\begin{equation} \label{eq:post-bs}
p(\hBs|\Db) = c \ p(\hBs) \ p(\Db|\hBs) =
  c \ p(\hBs) \int p(\Db|\TBs,\hBs) \ p (\TBs|\hBs) \ d \TBs
\end{equation}
where $c$ is a normalization constant.
We can then select a DAG model
that has a high posterior probability or average
several good models for prediction.  

The problem of selecting an appropriate DAG model,
or sets of DAG models, given data, posses
a serious computational challenge, 
because the number of DAG models grows faster than exponential
in $n$.  Methods for searching through the
space of model structures are discussed (e.g.) 
by Cooper and
Herskovits (1992), Heckerman, Geiger, and Chickering
(1995a), and Friedman and Goldszmidt (1997).
\nocite{CH92,HGC95ml,FG97}

From a statistical viewpoint, an important question which needs to be
addressed is how to specify the quantities $p(\hBs)$,
$p(\Db|\TBs,\hBs)$, $p (\TBs|\hBs)$, needed for evaluating
$p(\hBs|\Db)$ for every DAG model $\Bm$ that could conceivably be
considered by a search algorithm.  Buntine (1991) and Heckerman et
al. (1995a) discuss methods for specifying the priors $p(\hBs)$ via a
small number of direct assessments.

Herein, we develop practical methods for assigning 
parameter priors, $p (\TBs|\hBs)$, to every candidate DAG
model $\Bm$ via a small number of direct assessments. 
Our method is based
on a set of assumptions, the most notable of which 
is the assumption that complete DAG models represent
the same set of distributions, which implies that
data cannot distinguish between two complete DAG models.
Multivariate Gaussian, multinomial, and multivariate 
$t$ distributions satisfy this assumption.
Another assumption is
{\em likelihood and prior modularity}, 
which says that the local distribution for $x_i$
and its parameter priors
depend only on the parents of $x_i$ but
not on the entire description of the structure.
These assumptions, together with 
{\em global parameter independence}, introduced by
Spiegelhalter and Lauritzen (1990)\nocite{SL90},
are the heart of the proposed methodology.

The methodology described herein for setting
priors to DAG models and consequently calculating their 
marginal likelihoods is an extension of the
results by Dawid and Lauritzen (1993)\nocite{DL93}
for decomposable graphical models.
For decomposable graphical models, which is a set of models
that can be regarded both as DAG models 
as well as undirected graphical models,
the two methodologies are identical.
Our specification of a formal set of assumptions followed by
a technical derivation of this methodology 
provides an easy access to examine the validity
of the approach and devise alternatives when needed.

The contributions of this paper are as follows:
A methodology for specifying parameter priors for
many DAG structures using a few direct assessments
(Section~2).
A formula that computes the marginal likelihood
for every DAG model (Section~3). 
A specialization of this formula
to an efficient computation for
Gaussian DAG models (Section~4).
An analysis of complete Gaussian DAG models 
which shows that the only parameter prior
that satisfies our assumptions
is the normal-Wishart distribution (Section~5).
The analysis is based on the
following new characterization of the 
Wishart, normal, and normal-Wishart distributions.

\vspace{1ex}

\noindent
{\bf Theorem}
{\em 
Let $W$ be an $n \times n$, $n \ge 3$, positive-definite symmetric 
matrix of real random variables such that no entry in $W$ is zero,
$\mu$ be an $n$-dimensional vector of random variables,
$f_W(W)$ be a pdf of $W$,
$f_{\mu}(\mu)$ be a pdf of $\mu$,
and $f_{\mu,W}(\mu,W)$ be a pdf of $\{\mu,W\}$.
Then, 
$f_W(W)$ is a Wishart distribution,
$f_{\mu}(\mu)$ is a normal distribution,
and $f_{\mu,W}(\mu,W)$ is a normal-Wishart distribution
if and only if global parameter independence holds
for unknown $W$, unknown $\mu$, or unknown $\{\mu,W\}$,
respectively.
}

The assumption of global parameter independence
is expressed differently for each of the three cases
treated by this theorem and the 
proof follows from
Theorems~\ref{thm:Wishart},~\ref{thm:normal} 
and~\ref{thm:normalWishart}, respectively,
proven in Section~5.
It should be noted that a single principle, global parameter
independence,
is used to characterize three different distributions.

A similar characterization for the bivariate
Wishart, bivariate normal, and bivariate normal-Wishart 
distributions has recently been obtained 
under the assumption that the pdf is strictly positive,
and assuming also some additional independence 
constraints---termed standard local parameter independence
(Geiger and Heckerman, 1998).\nocite{GH98pms}
Another related result is the
characterization of the Dirichlet distribution
via global and local parameter independence
(Geiger and Heckerman, 1997; \Jarai, 1998). \nocite{GH97stat,Ja98}

\section{The Construction of Parameter Priors} 
\label{sec:like-pri}

In this section, we present assumptions that simplify the assessment
of parameter priors and a method of assessing these priors.
The assumptions are as follows:

\begin{assumption}[Complete model equivalence] \label{ass:cme}
Let $\Bmone = (\Bsone, \Fsone)$ be a complete DAG model
for $\U$.
The family $\Fstwo$ of every complete DAG model
$\Bmtwo = (\Bstwo, \Fstwo)$ for $\U$ is such that
$\Bmone$ and $\Bmtwo$ 
represent the same set of joint probability distributions,
namely, that for every 
$\TBsone$ there exists $\TBstwo$ such that
$p(\C | \TBsone,\hBsone) = p(\C | \TBstwo,\hBstwo)$ 
and vice versa.
\end{assumption}

Two examples where this assumption holds are quite common.
One happens when
$p(\C | \TBsone,\hBsone)$
and
$p(\C | \TBstwo,\hBstwo)$
are multivariate normal distributions
and the other happens when
$\U$ consists of variables with finite domains
and
$p(\C | \TBsone,\hBsone)$
and
$p(\C | \TBstwo,\hBstwo)$
are unrestricted discrete distributions.
In these two cases, all the local distributions
have the same functional form in every ordering of the 
variables. If the joint distribution for $\U$ is 
a multivariate $t$ distribution, then too,
all local conditional distributions
have the same functional form
(e.g., DeGroot, 1970)\nocite{DeGroot70},
however, unlike the unrestricted discrete
and multivariate normal distributions,
for $t$ distributions, the parameters 
of the local distributions are dependent which violates
assumption~\ref{ass:pi} discussed below.

We now provide an example where
this assumption fails.
Suppose the set of variables $\U=\{\X_1,\X_2,\X_3\}$ consists of 
three variables each with possible 
values $\{x_i, \overline{x}_i\}$, respectively, and
$\Bsone$ is the complete structure with arcs
$\X_1 \rightarrow \X_2$, $\X_1 \rightarrow \X_3$, and $\X_2
\rightarrow \X_3$. 
Suppose further, that
the local distributions $\Fsone$ of model $\Bmone$ 
are restricted to the logit
\begin{displaymath}
p(\x_i|\paSi,\Thi,\hBs)= \frac{1}{
  1 + {\rm exp}\left\{ a_i + \sum_{\x_j \in \paSi} b_{ji} \x_j \right\}}
\end{displaymath}
where $\theta_1= \{a_1\}$,
$\theta_2= \{a_2, b_{12}\}$,
and $\theta_3 = \{a_3, b_{13}, b_{23}\}$.

Consider now a second 
complete model $\Bmtwo$ for $\U=\{\X_1,\X_2,\X_3\}$
whose structure consists of the arcs $\X_1 \rightarrow \X_2$, 
$\X_1 \rightarrow \X_3$, and $\X_3 \rightarrow \X_2$.  
Assumption~\ref{ass:cme} asserts that the families of 
local distributions for $\Bmone$ and $\Bmtwo$ are such that
the set of joint distributions for $\U$ represented
by these two complete models is the same.  
In this example, however, if we specify the local families
for $\Bmtwo$ by also restricting
them to be logit distributions,
then the two models will represent different sets of
joint distributions over
$\{\X_1,\X_2,\X_3\}$. Hence,
Assumption~\ref{ass:cme} will be violated.
Using Bayes rule one can always determine a 
set of local distribution families that
will satisfy Assumption~\ref{ass:cme}, however,
their functional form will usually involve an integral
(and will often violate Assumption~\ref{ass:pi} below).

Note that whenever two DAG models represent the same set
of probability distributions for $\U$, they must also specify
the same set of independence assumptions. The example 
with the logit distributions highlights that
the converse does not hold because every complete DAG
represents the same independence assumptions, namely none, and
yet complete DAG models can represent
different sets of probability distributions.

Our definition of $\hBs$, that the true
joint pdf of a set of variables $\U$ is perfectly represented by $m$,
and Assumption~\ref{ass:cme}, which says that two complete
models represent the same set of joint pdfs for $\U$, 
imply that for two complete models $\hBsone = \hBstwo$.
This is a strong assumption.
It implies that 
$p(\TBstwo|\hBstwo) = p(\TBstwo|\hBsone)$
because two complete models represent the same set of distributions.  
It also implies $p(\Db|\hBsone) = p(\Db|\hBstwo)$ which says
that the marginal likelihood for two complete DAG models is
the same for every data set, or equivalently, that complete
DAG models cannot be distinguished by data.
Obviously, in the example with the logit distributions,
the two models can be distinguished by data 
because they do not represent
the same set of joint distributions.

\begin{assumption}[Regularity] \label{ass:regular}
For every two complete DAG models $\Bmone$ and $\Bmtwo$ for $\U$
there exists a one-to-one mapping $k_{1,2}$
between the parameters $\TBsone$ of $\Bmone$
and the parameters $\TBstwo$ of $\Bmtwo$
such that the likelihoods satisfy
$p(\C|\TBsone,\hBsone)  = p(\C|\TBstwo,\hBstwo) $
where $\TBstwo=k_{1,2}(\TBsone$).
The Jacobian
$|\dXdY{\TBsone}{\TBstwo}|$ exists and is 
non-zero for all values of $\CTBsone$.
\end{assumption}

\noindent 
Assumption~\ref{ass:regular} implies
$p(\TBstwo|\hBsone) =  \left| \frac{\partial \TBsone}{\partial \TBstwo}
\right| \ p(\TBsone|\hBsone)$ where
$\TBstwo=k_{1,2}(\TBsone$).
Furthermore, 
due to Assumption~\ref{ass:cme},
$p(\TBstwo|\hBstwo) = p(\TBstwo|\hBsone)$,
and thus
\begin{equation}
\label{prior_eq}
p(\TBstwo|\hBstwo) =  \left| \frac{\partial \TBsone}{\partial \TBstwo}
\right| \ p(\TBsone|\hBsone).
\end{equation}

For example, suppose $\x= \{x_1,x_2\}$ have a non-singular Bivariate
normal pdf $f(\C)= N(\C| \vecmu,W)$
where $\vecmu$ is the vector of means and $W = (w_{ij})$ 
is the inverse of
a positive definite covariance matrix.  If we write
$f(\C) = f_{x_1}(x_1) f_{x_2|x_1}(x_2 | x_1)$ where 
$f_{x_1}(x_1)= N(x_1 | e_1, 1/v_1)$ and 
$f_{x_2|x_1}(x_2 | x_1)= N(x_2 | e_{2|1}+ b_{12} x_1, 1/v_{2|1})$, then
the following well known relationships
are satisfied:
\begin{equation}
\label{basic1}
w_{11} = \frac{1}{v_1} + \frac{b_{12}^2}{v_{2|1}} \ \ \ \
w_{12} = - \frac{b_{12}}{v_{2|1}} \ \ \ \
w_{22} = \frac{1}{v_{2|1}} \ \ \ \
e_1 = \mu_1  \ \ \ \
e_{2|1} = \mu_2 - b_{12} \mu_1 \ \ \ \
\end{equation}
 Note that the
transformation between $\{\vecmu, W\}$ and
$\{e_1, v_1, e_{2|1}, v_{2|1}, b_{12}\}$ is one to
one and onto as long as $W$ is the inverse of a covariance
matrix and the conditional variances $v_1, v_{2|1}$ are positive.  
The Jacobian of this transformation is given by,
\begin{equation}
\label{eq:fulljac}
|\frac
{\partial{w_{11},w_{12},w_{22}, \mu_1,\mu_2}}
{\partial{v_1,v_{1|2},b_{12}, e_1,e_{2|1}}}
|
=
v_1^{-2} v_{2|1}^{-3}
\end{equation}
Symmetric equations hold when $f(\C)$ is written
as $f_{x_2}(x_2) f_{x_1|x_2}(x_1 | x_2)$ 
and so there is a one-to-one and onto mapping between
$\{e_1, v_1, e_{2|1}, v_{2|1}, b_{12}\}$ and
$\{e_2, v_2, e_{1|2}, v_{1|2}, b_{21}\}$.
Note that the parameters $\vecmu, W$
for the joint space are instrumental
for decomposing the needed mapping into a composition
of two mappings.

\begin{assumption}[Likelihood Modularity] \label{ass:lm}
For every two DAG models $\Bmone$ and $\Bmtwo$ for
$\U$ such that $\X_i$ has the same parents in $\Bmone$ and $\Bmtwo$,
the local distributions for $x_i$ in both models
are the same, namely,
$p(\x_i|\paSi,\Thi,\hBsone) = p(\x_i|\paSi,\Thi,\hBstwo)$
for all $\X_i \in \U$.
\end{assumption}

\begin{assumption}[Prior Modularity] \label{ass:pm}
For every two DAG models $\Bmone$ and $\Bmtwo$ for
$\U$ such that $\X_i$ has the same parents in $\Bmone$ and $\Bmtwo$,
$p(\Thi|\hBsone) = p(\Thi|\hBstwo)$.
\end{assumption}

\begin{assumption}[Global Parameter Independence] \label{ass:pi}
For every DAG model $\Bm$ for $\U$, 
$p(\TBs|\hBs) = \prod_{i=1}^n p(\Thi|\hBs)$.
\end{assumption}

The likelihood and prior modularity assumptions have been used
implicitly in the work of (e.g.) Cooper and Herskovits (1992),
Spiegelhalter, Dawid, Lauritzen, and Cowell (1993), and Buntine
(1994).\nocite{CH92,SDLC93,Buntine94} Heckerman et al.\
(1995a)\nocite{HGC95ml} made \Ass{ass:pm} explicit in the context of
discrete variables under the name parameter modularity.  Spiegelhalter
and Lauritzen (1990)\nocite{SL90} introduced Assumption~\ref{ass:pi}
in the context of DAG models under the name global independence.

Note that the first three assumptions concern the distribution of
$\bX$ whereas the last two assumptions concern the distribution of the
parameters.  Obviously, when the parameters $\theta_1, \ldots
\theta_n$ are not variation independent for every complete DAG model
for $\bf X$, the assumption of global parameter independence is
inconsistent with the model and can not be true.  Hence,
assumption~\ref{ass:pi} excludes, for example, the possibility that
two local distributions share a common parameter.  On the other hand,
even when the parameters are variation independent, it is possible to
specify a prior distribution for $\theta$ that violates global
parameter independence.  Cowell et al. (1999, pp.\ 191-2) highlight
this point.

The assumptions we have made lead to the following significant consequence:
When we specify a parameter prior $p(\TBsc|\hBsc)$ for one complete
DAG model $\Bmc$, we also implicitly specify a prior $p(\TBs|\hBs)$
for any DAG model $\Bm$ among the super exponentially many possible
DAG models.  Consequently, we have a framework in which a manageable
number of direct assessments leads to all the priors needed to search
the model space.  In the rest of this section, we explicate how all
parameter priors are determined by the one elicited prior.  In
Section~4, we show how to elicit the one needed prior $p(\TBsc|\hBsc)$
under specific distributional assumptions.

Due to the complete model equivalence and regularity assumptions, 
we can compute $p(\TBsc|\hBsc)$ for one complete model
for $\U$ from the
prior of another complete model for $\U$.  In so doing, 
we are merely performing coordinate transformations between
parameters for different variable orderings in the factorization of
the joint likelihood (Eq.~\ref{prior_eq}).
Thus by specifying parameter prior for
one complete model, we have implicitly specified a prior
for every complete model.

It remains to examine how the prior
$p(\TBs|\hBs)$ is computed
for an incomplete DAG model $\Bm$ for $\U$
from the prior $p(\TBsc|\hBsc)$ for some
complete model $\Bmc$.
Due to global parameter independence we have
$p(\TBs|\hBs) = \prod_{j=1}^n p(\Thj|\hBs)$
and therefore it suffices to examine each of the $n$ terms
separately.  To compute
$p(\Thi|\hBs)$, we identify a complete DAG model $\Bmci$
such that $\PaSi=\PaScii$.
The prior $p(\TBsci|\hBsci)$ is obtained from
$p(\TBsc|\hBsc)$, as we have shown for every
pair of complete DAG models.
Due to global parameter independence 
$p(\TBsci|\hBsci)$ is a product one
term of which is $p(\Thi|\hBsci)$.
Finally, due to prior modularity 
$p(\Thi|\hBs)$ is equal to $p(\Thi|\hBsci)$.

This methodology of constructing priors is described by
Heckerman et al.\ (1995a)
\nocite{HGC95ml}
for discrete DAG models and in Section~\ref{sec:linear} 
for Gaussian DAG models.
Our method is equivalent to the method of compatible priors
devised for decomposable graphical models 
(Dawid and Lauritzen, 1993). \nocite{DL93}
Our arguments, via a set of assumptions,
can be regarded as an axiomatic justification  
for compatible priors, and as an extension of this
method to general DAG models and to any probability distributions that
satisfy Assumptions~\ref{ass:cme} -- \ref{ass:pi}.
We are currently unaware, however, of additional
probability distributions that satisfy
these five assumptions.

The following theorem summarizes the general construction
which was formulated to cover both cases---the discrete
and the Gaussian.

\begin{theorem} \label{thm:prior}
Given Assumptions
\ref{ass:cme} through
\ref{ass:pi}, 
the parameter prior 
$p(\TBs|\hBs)$ for every DAG model $\Bm$ 
is determined by a specified parameter prior $p(\TBsc|\hBsc)$
for an arbitrary complete DAG model $\Bmc$.
\end{theorem}

Theorem~\ref{thm:prior} shows that once we specify the 
parameter prior for one complete DAG model all other priors
can be generated automatically and need not be specified manually.
Consequently, together with Eq.~\ref{eq:post-bs} and due
to the fact that also likelihoods can be generated
automatically in a similar fashion, we have 
a manageable methodology to automate the computation of
$p(\Db|\hBs)$ for any DAG model of $\U$ which is being
considered by a search algorithm as a candidate model.
Next we show how this computation can be done efficiently.

\section{Computation of the Marginal Likelihood for Complete Data} 
\label{sec:be}

For a given $\U$, consider a DAG model $\Bm$ and a complete random
sample $\Db$.  Assuming global parameter independence, the parameters
remain independent given complete data.  That is,
\begin{equation} \label{eq:post-pi} 
p(\TBs|\Db,\hBs) = 
  \prod_{i=1}^n p(\Thi|\Db,\hBs) 
\end{equation}
In addition, assuming global parameter independence, likelihood
modularity, and prior modularity, the parameters remain modular
given complete data.  In particular, if $\X_i$ has the same parents in
$\Bsone$ and $\Bstwo$, then 
\begin{equation} \label{eq:post-pm} 
p(\Thi|\Db,\hBsone) = p(\Thi|\Db,\hBstwo)
\end{equation}
Also, for any $\bY \subseteq \U$, define $\Db^{\bY}$ to be the random
sample $\Db$ restricted to observations of $\bY$.  For example, if
$\U=\{\X_1,\X_2,\X_3\}$, $\bY=\{\X_1,\X_2\}$, and
$\Db=\{\C_1=\{\x_{11},\x_{21},\x_{31}\},
\C_2=\{\x_{12},\x_{22},\x_{32}\}\}$, then we have $\Db^{\bY} = \{
\{\x_{11},\x_{21}\}, \{\x_{12},\x_{22}\}\}$.  
Let $\bY$ be a subset of $\U$, and $\Bsc$ be a complete structure for
any ordering where the variables in $\bY$ come first.  Then, assuming
global parameter independence and likelihood modularity, it is not
difficult to show that
\begin{equation} \label{eq:ignore}
p(\by|\Db,\hBsc) = p(\by|\Db^{\bY},\hBsc)
\end{equation}
Given these observations, we can compute the marginal likelihood as
follows, yielding an important 
component for searching
DAG models via a Bayesian methodology.

\begin{theorem} \label{thm:be}
Given any complete DAG model $\Bmc$ for $\U$, any DAG model $\Bm$ for
$\U$, and any complete random sample $\Db$, Assumptions~\ref{ass:cme}
through \ref{ass:pi} imply
\begin{equation} \label{eq:be}
p(\Db|\hBs) = \prod_{i=1}^n 
  \frac{p(\Db^{\Pai \cup \{\X_i\}}|\hBsc)}{
        p(\Db^{\Pai}|\hBsc)}
\end{equation}
\end{theorem}

\noindent {\bf Proof:}
From the rules of probability, we have

\begin{equation} \label{eq:be1a}
p(\Db|\hBs) = \prod_{l=1}^{\textcolor{blue}{N}} \int p(\C_l|\TBs,\hBs) \ p(\TBs|\Db_l,\hBs) 
  \ d\TBs
\end{equation}

\noindent
where $\Db_l = \{\C_1,\ldots,\C_{l-1}\}$.  Using
Equations~\ref{eq:bn-def-p} and \ref{eq:post-pi} to rewrite the first
and second terms in the integral, respectively, we obtain

\begin{displaymath} \label{eq:be1b}
p(\Db|\hBs) = \prod_{l=1}^{\textcolor{blue}{N}} \int \prod_{i=1}^n p(\x_{il}|\pail,\Thi,\hBs) \ 
  p(\Thi|\Db_l,\hBs) \ d\TBs
\end{displaymath}
where $\x_{il}$ is the value of $X_i$ in the $l$-th data point.

\noindent
Using likelihood modularity and \Eq{eq:post-pm}, we get

\begin{equation} \label{eq:be2}
p(\Db|\hBs) = \prod_{l=1}^{\textcolor{blue}{N}} \int \prod_{i=1}^n p(\x_{il}|\pail,\Thi,\hBsci) \ 
  p(\Thi|\Db_l,\hBsci) \ d\TBs
\end{equation}

\noindent
where $\Bsci$ is a complete structure with variable ordering
$\Pai$, $\X_i$ followed by the remaining variables.  Decomposing
the integral over $\TBs$ into integrals over the individual parameter
sets $\Thi$, and performing the integrations, we have
\begin{displaymath} \label{eq:be3}
p(\Db|\hBs) = \prod_{l=1}^{\textcolor{blue}{N}} \prod_{i=1}^n 
  p(\x_{il}|\pail,\Db_l,\hBsci)
\end{displaymath}
Using \Eq{eq:ignore}, we obtain
\begin{eqnarray} \label{eq:be4}
p(\Db|\hBs) & = & \prod_{l=1}^{\textcolor{blue}{N}} \prod_{i=1}^n 
  \frac{ p(\x_{il},\pail|\Db_l,\hBsci) }{ p(\pail|\Db_l,\hBsci) }
     \nonumber \\*[9pt]
& = & \prod_{l=1}^{\textcolor{blue}{N}} \prod_{i=1}^n 
  \frac{
    p(\x_{il},\pail|\Db^{\Pai \cup \{\X_i\}}_l,\hBsci) }{
     p(\pail|\Db^{\Pai}_l,\hBsci) }
     \nonumber \\*[9pt]
& = & \prod_{i=1}^n 
  \frac{
     p(\Db^{\Pai \cup \{\X_i\}}|\hBsci) }{
     p(\Db^{\Pai}|\hBsci) }
\end{eqnarray}
By the likelihood modularity, complete model equivalence,
and regularity assumptions, we have that
$p(\Db|\hBsci) = p(\Db|\hBsc), \ i=1,\ldots,n$.
Consequently, for any subset $\bY$ of $\U$, we obtain
$p(\Db^{\bY}|\hBsci) = p(\Db^{\bY}|\hBsc)$ by summing over
the variables in $\U \setminus \bY$.  
Consequently, using \Eq{eq:be4}, we get \Eq{eq:be}. \qed

An equivalent approach for computing the marginal likelihood
(\Eq{eq:be}) 
for decomposable discrete and Gaussian DAG models 
has been developed 
by Dawid and Lauritzen (1993)\nocite{DL93}
using compatible priors.

An important feature of \Eq{eq:be}, which we now demonstrate,
is that two DAG models that represent 
the same assertions of conditional independence have the same 
marginal likelihood.
We say that two structures for $\U$ are {\em independence
equivalent} if they represent the same assertions of conditional
independence.  Independence equivalence is an equivalence relation,
and induces a set of equivalence classes over the possible structures
for $\U$.  

Verma and Pearl (1990)\nocite{Verma90} provide a simple
characterization of independence equivalent structures
using the concept of a v-structure.
Given a
structure $\Bs$, a {\em v-structure} in $\Bs$ is an ordered node
triple $(\X_i,\X_j,\X_k)$ where $\Bs$ contains the arcs $\X_i
\rightarrow \X_j$ and $\X_j \leftarrow \X_k$, and there is no arc
between $\X_i$ and $\X_k$ in either direction.
Verma and Pearl show that
two structures for $\U$ are independence equivalent if and
only if they have identical edges and identical v-structures.
This characterization makes it easy to identify independence
equivalent structures.  

An alternative characterization  developed
by Chickering (1995) \nocite{Ci95}
and independently by
Andersson, Madigan, and Perlman (1997, Lemma 3.2),
\nocite{AMP97}
is useful for proving our claim 
that independence equivalent structures
have the same marginal likelihood.  
An {\em arc reversal} is a transformation from one
structure to another, in which a single arc between two nodes is
reversed.  An arc between two nodes is said to be {\em covered\/} if
those two nodes would have the same parents if the arc were
removed.
\begin{theorem}[Chickering, 1995; Andersson, Madigan, and Perlman, 1997]
\label{thm:C95}
\nocite{AMP97}
Two structures for $\U$ are independence equivalent if and only if
there exists a set of covered arc reversals that transform one
structure into the other.
\end{theorem}
A proof of this theorem can also be found in Heckerman et~al. (1995a).

Theorem~\ref{thm:C95} implies that
if every pair of DAGs that differ by a single covered arc 
represent the same
set of distributions, then every two independence equivalent
DAGs represent the same set of distributions.
Furthermore, a consequence of the next theorem is that
Assumptions~\ref{ass:cme} through \ref{ass:pi}
imply that indeed every two independence equivalent
DAGs represent the same set of distributions.
Without these assumptions, two independence equivalent
DAGs can represent different sets of distributions.

\begin{theorem} \label{thm:samelikelihood}
Given Assumptions~\ref{ass:cme} through \ref{ass:pi},
every two independence equivalent 
DAG models have the same marginal likelihood.
\end{theorem}

\noindent {\bf Proof:} 
Theorem~\ref{thm:C95} implies that we can restrict 
the proof to two DAG models that differ 
by a single covered arc. Say the arc is between $X_i$
and $X_j$ and that the joint parents of $X_i$ and $X_j$
are denoted by $\pi$.
For these two models, \Eq{eq:be} differs only
in terms $i$ and $j$. For both models
the product of these terms is 
$ p(\Db^{\pi \cup \{\X_i,X_j\}}|\hBsc)/ 
p(\Db^{\pi}|\hBsc)$.
\qed

The conclusions of Theorem~\ref{thm:be},
and, consequently, of Theorem~\ref{thm:samelikelihood} 
are not justified when our assumptions are violated. 
In the example of the logit distributions, discussed in
the previous subsection, which violates assumption~\ref{ass:cme},
the structures $\Bsone$ and $\Bstwo$ differ by the
reversal of a covered arc between $\X_2$ and $\X_3$, but, given 
that all local distribution families are
logit, there are certain joint distributions that can
be represented by one structure, but not the other,
and so their marginal likelihood will be different.

The implication of Theorem~\ref{thm:samelikelihood} 
is quite strong: all models in
the same independence equivalence class are scored equivalently.
This severely constrains possible parameter priors
as shown in the next two sections.
One possible approach to bypass our assumptions
is to select one representative DAG
model from each class of independence 
equivalent DAG models, assume global
parameter independence only for these representatives,
and evaluate the marginal likelihood only
for these representatives. 
The search can then be conducted in the space of
representative models as suggested in 
Spirtes and Meek (1995), Chickering (1996), and
Madigan, Andersson, Perlman, and Volinsky (1996)\nocite{SM95,MPV96,Ch96}.
The difficulty with this approach is that
when projecting a prior from a complete DAG model
to a DAG model with missing edges, one needs to perform
additional high dimensional integrations before using the parameter
modularity property (see Section~2). 
Another approach is to modify the definition
of $\hBs$ to allow independence equivalent DAG models to have
different parameter priors.
This alternative is needed when arcs have a causal interpretation.
However, when choosing this alternative, the parameter prior
for each model examined by a search procedure
must be provided by a user as the search is being
conducted, or a new mechanism to 
produce acceptable priors on-the-fly must be devised.

\section{Gaussian Directed Acyclic Graphical Models}
\label{sec:linear} 

We now apply the methodology
of previous sections to Gaussian DAG models.
A Gaussian DAG model is a DAG model as defined
by Eq~\ref{eq:bn-def-p}, where 
each variable $\X_i \in \U$ is continuous, and
each local likelihood is the linear regression model
\begin{equation} \label{eq:norm-i}
p(\x_i|\paSi,\Thi,\hBs)=
  N(\x_i|m_i + \sum_{\x_j \in \pai} b_{ji} \x_j, 1/v_i)
\end{equation}
where $N(\x_i|\mu,\tau)$ is a normal distribution with mean $\mu$ and
precision $\tau>0$.  Given this form, a missing arc from $\X_j$ to
$\X_i$ 
is equivalent to $b_{ji}=0$ in the DAG model.  The
local parameters are given by $\Thi=(m_i,\vecb_i,v_i)$, where
$\vecb_i$ is the column vector $(b_{1i}, \ldots, b_{i-1,i})$
of regression coefficients. Furthermore, $m_i$ is the conditional mean
of $X_i$ and $v_i$ is the conditional variance of $X_i$.

For Gaussian DAG models, the joint likelihood
$p(\C|\TBs,\hBs)$ obtained from Eqs~\ref{eq:bn-def-p}
and~\ref{eq:norm-i}
is an $n$-dimensional multivariate normal distribution
with a mean vector $\vecmu$ and a symmetric positive definite 
precision matrix $\matW$,
\[
p(\C|\TBs,\hBs) = \prod_{i=1}^n p(\x_i|\paSi,\Thi,\hBs)
= N(\C|\vecmu, \matW).
\]

For a complete model $\Bmc$ with ordering $(\X_1,\ldots,\X_n)$ 
there is a one-to-one mapping between $\TBsc = \bigcup_{i=1}^n \Thi$ 
where $\Thi=(m_i,\vecb_i,v_i)$ 
and $\{\vecmu,\matW\}$ which has a nowhere singular Jacobian matrix.
Consequently, assigning a prior for the parameters 
of one complete model induces a parameter prior, via the 
change of variables formula, for
$\{\vecmu,\matW\}$ and in turn, induces
a parameter prior for every complete model.
Any such induced parameter prior must satisfy,
according to our assumptions, global parameter independence.
Not many prior distributions satisfy such a requirement.  
In fact, in the next section we show
that the parameter prior $p(\vecmu,\matW|\hBsc)$ 
must be a normal-Wishart distribution.

For now we proceed by simply choosing $p(\vecmu,\matW|\hBsc)$
to be a normal-Wishart distribution.  In particular,
$p(\vecmu|\matW,\hBsc)$ is a multivariate normal distribution
with mean $\nu$ and precision matrix $\amu
\matW$ ($\amu > 0$); and $p(\matW|\hBsc)$ is a Wishart
distribution, given by,
\begin{equation} \label{eq:cln} 
p(\matW|\hBsc)=
  c(n,\aw) |T|^{\aw/2} |\matW|^{(\aw-n-1)/2} e^{-1/2\tr\{T \matW\}}  
 \equiv \W(\matW| \aw,T)
\end{equation}
with $\aw$ degrees of freedom $(\aw > n-1)$ and
a positive-definite precision matrix $T$
and where $c(n,\aw)$ is a normalization constant
given by
\begin{equation} 
\label{eq:normalization-constant}
c(n,\aw) = 
\left[ 2^{\aw n/2} \pi^{n(n-1)/4}
\prod_{i=1}^n \Gamma\left(\frac{\aw +1-i}{2}\right)\right]^{-1}
\end{equation} 
(DeGroot, 1970, pp.\ 57)\nocite{DeGroot70}.  We provide
interpretations for $\nu$, $\amu$, $T$, and $\aw$
later in this section.
Note that in some expositions of the Wishart distribution, the inverse
of $T$ is used for the parameterization; $T^{-1}$ is called the scale
matrix (e.g., Press 1971, pp.\ 101).

This choice of a prior satisfies global parameter independence
due to the following well known theorem.

\textcolor{blue}{Let $M'$ denote the transpose of matrix $M$.}
Define a block partitioning $\{W_{11}, W_{12}, W'_{12}, W_{22}\}$ 
of an $n$ by $n$ matrix $W$ to be {\em compatible} with
a partitioning $\mu_1, \mu_2$ of an $n$ dimensional vector $\mu$,
if 
the indices of the rows that correspond to block $W_{11}$
are the same as the indices of the terms that constitute $\mu_1$
and similarly for $W_{22}$ and $\mu_2$.
Also define
\textcolor{blue}{$W_{11.2} 
= ((W^{-1})_{11})^{-1} = W_{11} - W_{12} W_{22}^{-1} W'_{12}$.}

\begin{theorem} \label{thm:pinw} 
If $f(\mu,W)$ is 
an $n$ dimensional normal-Wishart distribution, $n \geq 2$,
with parameters $\nu, \amu$, $\aw$, and $T$,
then $\{\mu_1, W_{11} - W_{12} W_{22}^{-1} W'_{12} \}$ 
is independent of 
$\{\mu_2 + W_{22}^{-1} W'_{12} \mu_1,$
$ W_{12},W_{22}\}$
for every partitioning $\mu_1, \mu_2$ of $\mu$ where
$W_{11}$,$W_{12}$, $W'_{12}$, $W_{22}$ is 
a block partitioning of $W$
compatible with the partitioning $\mu_1,\mu_2$.  
Furthermore, the pdf of
$\{\mu_1, W_{11.2}  \}$ 
is normal-Wishart with parameters
$\nu_1$, $\amu$, $T_{11}$,
and $\aw -n +l$ where 
$T_{11}$,$T_{12}$, $T'_{12}$, $T_{22}$ is 
a compatible block partitioning of $T$,
$\nu_1, \nu_2$ is a compatible partitioning of $\nu$, and
$l$ is the size of the vector $\nu_1$.
\end{theorem}

The proof of Theorem~\ref{thm:pinw} 
requires a change of variables from
$(\mu,W)$ to $(\mu_1$, $\mu_2 + W_{22}^{-1} W'_{12} \mu_1)$
and $(W_{11} - W_{12} W_{22}^{-1} W'_{12},$ $W_{12},W_{22})$.
Press (1971, p.\ 117-119)\nocite{Press71}
carries out these computations for the Wishart distribution.
Standard changes are needed
to obtain the claim for the normal-Wishart distribution.
A consequence of Theorem~\ref{thm:pinw}
is the following.

\begin{corollary}
\label{lem:wishart-consequence}
Let $W$ be a $n \times n$ positive-definite matrix 
of random variables.  Let $a$, $b$, and $c$ be three sets
of indices of $W$. If 
$f(W_{ab.c}) = \W(W_{ab.c} | \alpha_1,T_1)$
and
$f(W_{bc.a}) = \W(W_{bc.a} | \alpha_2,T_2)$,
then $\alpha_1 - l_{ab}  = \alpha_2 - l_{bc}$ 
where $l_{ab}$ is the number of indices in the block $a,b$
and $l_{bc}$ is the number of indices in the block $b, c$.
\end{corollary}
 
\noindent {\bf Proof:} 
The pdf for
$W_{b.ac} = (W_{ab.c})_{b.a} = (W_{cb.a})_{b.c}$ 
is a Wishart distribution, and from the two alternative ways
by which this pdf can be formed, 
using Theorem~\ref{thm:pinw}, 
it follows that $\alpha_1 - l_{ab} = \alpha_2 - l_{bc}$.
\qed

To see why the independence conditions
in Theorem~\ref{thm:pinw} imply global parameter
independence, consider the partitioning in which 
the first block contains the first $n-1$
coordinates which correspond to $X_1,\ldots, X_{n-1}$ 
while the second block contains the last
coordinate which corresponds to $X_n$.
For this partitioning, $b_n = - W_{22}^{-1} W'_{12}$,
$v_n =  W_{22}^{-1}$, and $m_n = \mu_2 + W_{22}^{-1} W'_{12} \mu_1$.
Furthermore, 
$((W^{-1})_{11})^{-1} = W_{11} - W_{12} W_{22}^{-1} W'_{12} = W_{11.2}$
is the precision matrix associated with $X_1,\ldots, X_{n-1}$.
Consequently, $\{m_n, b_n,v_n\}$
is independent of $\{\mu_1,W_{11.2}\}$.
We now recursively repeat this argument with
$\{\mu_1, W_{11.2} \}$ instead of $\{\mu, W\}$,
to obtain global parameter independence.
The converse, namely that global parameter independence 
implies the independence conditions in Theorem~\ref{thm:pinw},
is established similarly.

Our choice of prior implies that
the posterior $p(\vecmu,\matW|\Db,\hBsc)$ is also a normal-Wishart
distribution (DeGroot, 1970, p.\ 178)\nocite{DeGroot70}.  
In particular, 
$p(\vecmu|\matW,\Db,\hBsc)$,
where $\Db$ is a sample of $\m$ complete cases,
is multivariate normal with mean vector
given by
\begin{equation} 
\label{eq:updatemeans} 
\frac{\amu \nu + \m \, \Xvecbar_{\m}}{\amu+\m}
\end{equation} 
and precision matrix $(\amu+\m)\matW$, where $\Xvecbar_{\m}$ is the sample
mean of $\Db$, and $p(\matW|\Db,\hBsc)$ is a Wishart distribution with
$\aw+\m$ degrees of freedom and precision matrix $R$ given by
\begin{equation} \label{eq:updateTau} 
R =  
T + \matS_{\m} + \frac{\amu \m}{\amu+ \m}
  (\nu -\Xvecbar_{\m})(\nu -\Xvecbar_{\m})' 
\end{equation} 
where 
$\matS_{\m}= \sum_{l=1}^{N} (\Xvecsubl -\Xvecbar_{\m})
(\Xvecsubl -\Xvecbar_{\m})'$.
From these equations, we see
that $\amu$ and $\aw$ can be thought 
of as effective sample sizes for 
\textcolor{blue}{the normal and Wishart components of the prior}, respectively.

In order to calculate the marginal likelihood of a Gaussian DAG
model, we can work in the parametric space $(\vecmu, \matW)$. 
According to Theorem~\ref{thm:pinw},
if $p(\vecmu,\matW|\hBsc)$ is a 
normal-Wishart distribution with the parameters
given by the theorem, 
then $p(\vecmu_{\bY},((\matW^{-1})_{\bY \bY})^{-1}|\hBsc)$ is
also a normal--Wishart distribution with
effective sample sizes 
$\amu$ and $\aw-n+l$, 
and parameters 
$\nu_{\bY}$ and  
\textcolor{red}{$T_{\bY\bY}$},
where $\bY$ is a subset of $l$ coordinates 
and $\matM_{\bY\bY}$ is the matrix
$\matM$ with elements restricted to the corresponding variables in $\bY$.
Thus, we obtain the terms in \Eq{eq:be}:
\begin{equation} \label{eq:bge}
p(\Db^{\bY}|\hBsc)  = 
  (2\pi)^{-l \m /2} \left(\frac{\amu}{\amu+\m}\right)^{l/2}
  \frac{c(l,\awY)}{c(l,\awY+\m)} \ 
  |\textcolor{red}{T_{\bY\bY}}|^{\frac{\awY}{2}} \ 
  |\textcolor{red}{R_{\bY\bY}}|^{-\frac{\awY+\m}{2}} 
  \nonumber
\end{equation}
\textcolor{blue}{(See Geiger and Heckerman, 1994, for a derivation
when $l=n$.)}
Equations~\ref{eq:be}
and~\ref{eq:bge} together provide a way
to compute the marginal likelihood
for Gaussian DAG models given the assessment 
of the parameter prior $p(\vecmu,\matW|\hBsc)$.

\textcolor{blue}{Rather than a direct assessment of the parmeter prior,
 we consider
a partially indirect approach.  We start with the observation that when
$p(\vecmu,\matW|\hBsc)$ is normal--Wishart as we have described, then
then $p(\C|\hBsc)$ is a multivariate $t$ distribution with $\aw-n+1$
degrees of freedom, location vector $\nu$, and precision matrix
$\amu(\aw-n+1)/(\amu+1)T^{-1}$. 
This result can be derived by first integrating over $\vecmu$ using
Equation 6 on p.\ 178 of DeGroot with sample size equal to one, and then
integrating over $\matW$
following an approach similar to that on pp.\ 179--180. Next, when
$\aw > n+1$, it follows that
\begin{equation} \label{eq:t1}
{\rm E}(\bx|\hBsc) = \nu \ \ \ \ \ \ 
{\rm Cov}(\bx|\hBsc) = \frac{\amu+1}{\amu} \ \frac{1}{\aw-n-1} \ T
\end{equation}
(e.g., DeGroot, 1970, pp.\ 61).  
Thus, a person can assess the parameter prior
by assessing $\amu$ and $\aw$, driectly, and by assessing 
a DAG model for E$(\bx|\hBsc)$ and Cov$(\bx|\hBsc)$ and
then computing $\nu$ and $T$ using Equations~\ref{eq:t1}.
We call this model a {\em prior} DAG model.
The unusual aspect of this assessment is the conditioning hypothesis
$\hBsc$ (see Heckerman et al. [1995b] for a discussion).
This indirect approach provides a suitable Bayesian
alternative for many of the examples discussed in Spirtes, Glymour,
and Scheines (2001).\nocite{SGS01}}

\comment{
An alternative approach to
the assessment of the vector $\vecmu_0$, the matrix $T$, and the
quantities $\amu>0$, and $\aw>n-1$ is as follows.  From
Equations~\ref{eq:updatemeans} and \ref{eq:updateTau} and the
surrounding text, we see that $\nu$ and $T$ can be thought of as an
initial mean and scatter matrix, respectively, and that $\amu$ and
$\aw$ can be thought of as effective sample sizes for
them. Consequently, we can assess the effective sample sizes
directly, and then assess a {\em prior DAG mnodel}, infer the mean and
covariance of $\U$, $\vecmu_0$ and ${\rm Cov}(\U)_0$, respectively, and set

\begin{equation} \label{eq:t2}
\nu = \vecmu_0 \ \ \ \ \ \ T = \aw {\rm Cov}(\U)_0.
\end{equation} 

\noindent This method provides a suitable Bayesian
alternative for many of the examples discussed in Spirtes, Glymour,
and Scheines (2001).\nocite{SGS01}
}

\section{Characterization of Several Probability Distributions}

We now characterize the Wishart distribution 
as the only pdf that satisfies global parameter independence
for an unknown precision matrix $W$ with $n \geq 3$ coordinates
(Theorem~\ref{thm:Wishart}).
This theorem is phrased and proven in a
terminology that relates to known facts about the
Wishart distribution.
We proceed with similar characterizations of the
normal and normal-Wishart distributions
(Theorems~\ref{thm:normal} and~\ref{thm:normalWishart}).

We will use $\tr\{A+B\}$ to denote the sum of traces 
$\tr\{A\} + \tr\{B\}$ even when the dimensions of the
square matrices $A$ and $B$ are different.

\begin{theorem} \label{thm:Wishart}
Let $W$ be an $n \times n$, $n \ge 3$, positive-definite symmetric 
matrix of random variables
and $f(W)$ be a pdf of $W$.
Then, f$(W)$ is a Wishart distribution if and only if
$W_{11} - W_{12} W_{22}^{-1} W'_{12}$ is independent of 
$\{W_{12},W_{22}\}$
for every block partitioning $W_{11},W_{12}, W'_{12}, W_{22}$ of
$W$.
\end{theorem} 

\noindent {\bf Proof:} 
That
$W_{11.2}= W_{11} - W_{12} W_{22}^{-1} W'_{12}$ is independent 
of $\{W_{12},W_{22}\}$
whenever $f(W)$ is a Wishart distribution is a well
known fact (Press 1971, p.\ 117-119)\nocite{Press71}. 
It is also expressed by Theorem~\ref{thm:pinw}.
The other direction is proven by induction on $n$.
The base case $n=3$ is treated at the end.

The pdf of $W$ can be written in $n!$ orderings.
In particular, due to the assumed independence conditions,
and since the transformations
from $\{\A,\B,\D \}$ to
$\{W_{11.2},\B,\D \}$ and to
$\{W_{22.1},\A,\B \}$ both have a Jacobian 
determinant of 1,
we obtain the following functional equation:
{\small
\begin{equation}
\label{eq:wish1}
f(W) = f_1(\A - \B \D^{-1} \B') f_{2 \vb 1}(\D,\B) 
= f_2(\D - \B' \A^{-1} \B) f_{1 \vb 2}(\A,\B) 
\end{equation}}
where a subscripted $f$ denotes a pdf.

We divide the indices of $W$
into two blocks, the first block  (say, block 1)
contains $n-1$ indices
and the second block (say, block 2) consists of one index.
By the induction hypothesis, and since the independence conditions
on $W$ also hold for $W_{11.2}$
we conclude that
$W_{11.2}$ is distributed according to 
$\W(W_{11.2}|\; \alpha_1,T_{1})$.
Since this argument holds for every
block of size $n-1$ of $W$, and since if a matrix $V$
is distributed Wishart so does
$V_{11.2}$ for any block of indices
(Theorem~\ref{thm:pinw}),
it follows that 
$W_{11.2}$ is distributed according to 
$\W(W_{11.2}|\; \alpha_1,T_{1})$
also for blocks of size smaller than $n-1$.

Thus,
\begin{eqnarray}
\label{eq:wish2}
\nonumber
\lefteqn{
c_1 |\A-\B\D^{-1}\B'|^{\beta_1}
e^{\tr\{T_{1} (\A-\B\D^{-1}\B')\}} f_{2 \vb 1}(\D,\B)  = }
\\
\;\;\;
& & 
c_2 |\D - \B'\A^{-1} \B |^{\beta_2} 
e^{\tr\{T_{2} (\D - \B' \A^{-1} \B ) \}} f_{1 \vb 2}(\A,\B) 
\end{eqnarray}
where $c_1$ and $c_2$ are normalizing constants.

We now argue that $\beta_1 = \beta_2$.
Divide the indices of $W$ to three non empty sets $a,b,c$
such that block 1 consists of the indices in $a,b$ 
and block 2 consists of the indices in $c$.
The matrices $W_{ab.c}$ and $W_{bc.a}$ have a Wishart distribution,
with, say, degrees of freedom $\alpha_1$ and $\alpha_2$ 
respectively,
and so according to Corollary~\ref{lem:wishart-consequence},
$\alpha_1 - l_{ab}  = \alpha_2 - l_{bc}$.
Furthermore, $W_{c.ab} = (W_{bc.a})_{c.b}$ 
has a Wishart distribution
with $\alpha_2 -l_{bc} + l_c$ degrees of freedom.
Consequently, $\beta_1 =  (\alpha_1 - l_{ab} -1 )/2$
is equal to $\beta_2 = ( \alpha_2 - l_{bc} + l_c - l_c -1)/2$.
Let $\beta =  \beta_1 = \beta_2$.

Define
\begin{eqnarray}
\label{eq:wish3}
F_{2 \vb 1}(\D,\B) & = & c_1 f_{2 \vb 1}(\D,\B) / |\D|^{\beta}
e^{\tr\{T_{2} \D + T_{1} (\B\D^{-1}\B')\}}
\\
\label{eq:wish3.1}
F_{1 \vb 2}(\A,\B) & = & c_2 f_{1 \vb 2}(\A,\B) / |\A|^{\beta}
e^{\tr\{T_{1} \A + T_{2} (\B' \A^{-1} \B) \}},
\end{eqnarray}
substitute into Equation~\ref{eq:wish2}, and obtain,
using $|\A-\B\D^{-1}\B'| |\D| = |W|$, that
$F_{2 \vb 1}(\D,\B) = F_{1 \vb 2}(\A,\B)$.
Consequently, $F_{2 \vb 1}$ and $F_{1 \vb 2}$
are functions only of $\B$ and thus, using
Equation~\ref{eq:wish1}, we obtain
\begin{equation}
\label{eq:wish4}
f(W) = 
|W|^{\beta} e^{\tr\{T_{1} \A + T_{2} \D\}} H(\B)
\end{equation}
for some function $H$.

To show that $f(W)$ is Wishart we must
find the form of $H$ and show that it is proportional
to  $e^{ 2 \tr \{ T_{12} W_{12}\}}$
for some matrix $T_{12}$.

Considering the three possible pairs of blocks 
formed with the sets of indices $a$, $b$, and $c$,
Equation~\ref{eq:wish4} can be rewritten as follows.
\begin{eqnarray}
\label{eq:wish6}
f(W) = 
|W|^{\beta_1} e^{\tr 
\{T_{aa} W_{aa} + T_{bb} W_{bb} + T_{cc} W_{cc} \}} 
e^{2 \tr \{T'_{ab} W_{ab} + T'_{ac} W_{ac} + T'_{bc} W_{bc} \}} 
H_1(W_{ac}, W_{bc})
\\
\label{eq:wish7}
f(W) = 
|W|^{\beta_2} e^{ \tr
\{S_{aa} W_{aa} + S_{bb} W_{bb} + S_{cc} W_{cc} \}} 
e^{2 \tr \{S'_{ab} W_{ab} + S'_{ac} W_{ac} + S'_{bc} W_{bc} \}} 
H_2(W_{ab}, W_{bc})
\\
\label{eq:wish8}
f(W) = 
|W|^{\beta_3} e^{ \tr
\{R_{aa} W_{aa} + R_{bb} W_{bb} + R_{cc} W_{cc} \}} 
e^{2 \tr \{R'_{ab} W_{ab} + R'_{ac} W_{ac} + R'_{bc} W_{bc} \}}
H_3(W_{ab}, W_{ac})
\end{eqnarray}
By setting 
$W_{ab}= W_{ac} = W_{bc}=0$, we get
$\beta_1= \beta_2 = \beta_3$ and
$T_{ii} = S_{ii} = R_{ii}$, for $i=a,b,c$.
By comparing Equations~\ref{eq:wish6} and~\ref{eq:wish7}
we obtain
\begin{equation}
\label{eq:wish9}
e^{2 \tr \{ (T'_{ac} - S'_{ac}) W_{ac}\}} 
H_1(W_{ac}, W_{bc})
=
e^{2 \tr \{
(S'_{ab} - T'_{ab}) W_{ab} + 
(S'_{bc} - T'_{bc}) W_{bc} \}} 
H_2(W_{ab}, W_{bc})
\end{equation}
Each side of this equation must be a function
only of $W_{bc}$. We denote this function by $H_{12}$.
Hence,
\[
H_1(W_{ac}, W_{bc}) = 
H_{12}(W_{bc})
e^{2 \tr \{ (S'_{ac} - T'_{ac}) W_{ac}\}} 
\]
and by symmetric arguments,
comparing Equations~\ref{eq:wish6} and~\ref{eq:wish8},
\[
H_1(W_{ac}, W_{bc}) = 
H_{13}(W_{ac})
e^{2 \tr \{ (R'_{bc} - T'_{bc}) W_{bc}\}} 
\]
Consequently,
$H_{12}(W_{bc})$ is proportional 
to  $e^{2 \tr \{ (R'_{bc} - T'_{bc}) W_{bc}\}}$
and so, substituting into Equation~\ref{eq:wish4},
$f(W)$ is found to be a Wishart distribution, as claimed.

It remains to examine the case $n=3$.  
We first assume $n=2$ in which case
$f(W)$ is not necessarily a Wishart distribution.
In the appendix we show that given the independence conditions for
two coordinates, $f$ must have the form
\begin{equation}
\label{eq:basis1}
f(W) = c |W|^{\beta} e^{\tr\{T W\}} H(\B)
\end{equation}
where $H$ is an arbitrary function,
and that the marginal distributions of
$W_{11.2}$ and $W_{22.1}$
are one dimensional Wishart distributions.

We now treat the case $n=3$ using 
these assertions about the case $n=2$.
Starting with Equation~\ref{eq:wish1},
and proceeding with blocks $a,b,c$ each containing exactly 
one coordinate,
we get, due to the given independence conditions for
two coordinates, that $f_1$ has the form
given by Equation~\ref{eq:basis1}, and that
$f_2$ is a one dimensional Wishart distribution.
Proceeding parallel to Equations~\ref{eq:wish2}
through~\ref{eq:wish3.1}, we obtain,

\begin{equation}
\label{eq:basis2}
H(a_{12} - b_1 b_2/W_{22}) F_{2 \vb 1}(\D,\B)
= F_{1 \vb 2}(\A,\B)
\end{equation}
where
$(b_1,b_2)$ is the matrix $W_{12}$, 
$a_{12}$ is the off-diagonal element of $W_{11}$,
$a_{12} - b_1 b_2/W_{22}$ is the off diagonal element
of $W_{11}- W_{12} W^{-1}_{22} W'_{12}$,
and $W_{22}$ is a $1\times 1$ matrix.
Note that the left hand side depends on $W_{11}$ only
through $a_{12}$. Thus also the right hand side
depends on $W_{11}$ only through $a_{12}$.  Let $b_1$ and $b_2$ be 
fixed, $y= b_1 b_2/W_{22}$, and $x= a_{12}$.  Also let
$F(t)= F_{2 \vb 1}(b_1 b_2/t,(b_1 ,b_2))$
and
$G(a_{12}) = F_{1 \vb 2}(\A,(b_1,b_2))$.
We can now rewrite Equation~\ref{eq:basis2} as
$H(x-y) F(y) = G(x)$.  
Now set $z=x-y$, and obtain
for every $y$ and $z$
\begin{equation}
\label{eq:basis5}
H(z) F(y) = G(y+z)
\end{equation}
the only measurable solution of which for $H$
is $H(z) = c e^{bz}$ (e.g., \Aczel, 1966)\nocite{Ac66}.

Substituting this form of $H$
into Equation~\ref{eq:basis1},
we see that $W_{11.2}$ has a two
dimensional Wishart distribution.
Recall that $W_{22.1}$ has a one dimensional
Wishart distribution.  Consequently, we can apply the
induction step starting form Equation~\ref{eq:wish2}
and prove the Theorem for $n=3$.  \qed

We now treat the situation when only the means are unknown,
characterizing the normal distribution.
The two-dimensional case turns out to be
covered by the
Skitovich-Darmois theorem 
(e.g., Kagan, Linnik, and Rao, 1973).
\nocite{KLR73}

\begin{theorem}[Skitovich-Darmois] \label{thm:Skitovich-Darmois}
Let $z_1,\ldots,z_k$ be independent random variables and
$\alpha_i, \beta_i$, $1<i<k$, be constant coefficients.
If $L_1 = \sum \alpha_i z_i$
is independent of $L_2 = \sum \beta_i z_i$,
then each $z_i$ for which $\alpha_i \beta_i \neq 0$
is normal.
\end{theorem}

The Skitovich-Darmois theorem is used in the proof of the base case of
our next characterization.  Several generalizations of the
Skitovich-Darmois theorem are described by Kagan et al.
(1973).\nocite{KLR73}

\begin{theorem} \label{thm:normal}
Let $W$ be an $n \times n$, $n \ge 2$, positive-definite symmetric 
matrix 
such that no entry in $W$ is zero,
$\mu$ be an $n$-dimensional vector of random variables,
and $f(\mu)$ be a pdf of $\mu$.
Then, f$(\mu)$ is an $n$ dimensional normal distribution 
$N(\mu | \eta, \gamma W)$ where $\gamma >0$ if and only if
$\mu_1$ is independent of $\mu_2 + W_{22}^{-1} W'_{12} \mu_1$
for every partitioning $\mu_1, \mu_2$ of $\mu$ where
$W_{11}$,$W_{12}$, $W'_{12}$, $W_{22}$ is a 
block partitioning of $W$ compatible
with the partitioning $\mu_1, \mu_2$.

\end{theorem}

\noindent {\bf Proof:} 
The two independence conditions,
$\mu_1$ independent of $\mu_2 + W_{22}^{-1} W'_{12} \mu_1$
and 
$\mu_2$ independent of $\mu_1 + W_{11}^{-1} W_{12} \mu_2$,
are equivalent to the following functional equation
\begin{equation}
\label{eq:normal1}
f(\mu) =
f_1(\mu_1) f_{2 \vb 1}( \mu_2 + W_{22}^{-1} W'_{12} \mu_1)
= 
f_2(\mu_2) f_{1 \vb 2}( \mu_1 + W_{11}^{-1} W_{12} \mu_2)
\end{equation}
where a subscripted $f$ denotes a pdf.
We show that the only solution for $f$ that satisfies
this equation is the normal distribution.
Consequently, both the if and only if portions
of the theorem will be established.

For $n \geq 3$, we can divide the indices of $W$
into three non-empty sets $a,b$ and $c$.
We group $a$ and $b$ to form a block and $b$ and $c$ to form a block.
For each of the two cases, let $W_{11}$ be the block 
consisting of the indices in $\{a,b\}$ or $\{b,c\}$, respectively,
and $W_{22}$ be the block consisting
of the indices of $c$ or $a$, respectively.
By the induction hypothesis applied to both cases
and marginalization we can assume that
$f_1(\mu_1)$ is a normal distribution
$N(\mu_1 | \eta_1, \gamma_1 ((W^{-1})_{11})^{-1})$ 
and that
$f_2(\mu_2)= 
N(\mu_2 | \eta_2, \gamma_2 ((W^{-1})_{22})^{-1})$.
Consequently, the pdf of 
the block corresponding to the indices in $b$ is
a normal distribution, and from the two alternative ways
by which this pdf can be formed, 
it follows that $\gamma_1 = \gamma_2$.

Let $\gamma = \gamma_i$, $i=1,2$, and define
\begin{eqnarray}
\nonumber
\label{eq:normal3}
F_{2 \vb 1}( x ) & = & 
f_{2 \vb 1}( x ) / 
N( x | \eta_2 +  W_{22}^{-1} W'_{12} \eta_1,  \gamma W_{22})
\\
\nonumber
\label{eq:normal4}
F_{1 \vb 2}( x ) & = & 
f_{1 \vb 2}( x ) /
N( x | \eta_1 + W_{11}^{-1} W_{12} \eta_2,  \gamma W_{11}).
\end{eqnarray}
By substituting these definitions into Equation~\ref{eq:normal1},
substituting the normal form for 
$f_1(\mu_1)$ and $f_2(\mu_2)$, 
and canceling 
on both sides of the equation 
the term $N(\mu|\eta,\gamma W)$ 
(which is formed by standard algebra pertaining 
to quadratic forms (E.g., DeGroot, p.\ 55)),
we obtain a new functional equation, 
\[
F_{2 \vb 1}( \mu_2 + W_{22}^{-1} W'_{12} \mu_1)
=
F_{1 \vb 2}( \mu_1 + W_{11}^{-1} W_{12} \mu_2).
\]
By setting $\mu_2 = - W_{22}^{-1} W'_{12} \mu_1$, we 
obtain
$F_{1 \vb 2}( (I - (W_{11}^{-1} W_{12})(W_{22}^{-1} W'_{12})) \mu_1)=
F_{2 \vb 1}( 0 )$ for every $\mu_1$.
Hence, the only solution to this functional equation is
$F_{1 \vb 2} =F_{2 \vb 1} \equiv \mbox{constant}$.
Consequently, $f(\mu) = N(\mu|\eta, \gamma W)$.

It remains to prove the theorem for $n=2$.
Let 
$z_1 = \mu_1$,
$z_2 = \mu_2 + w_{22}^{-1} w_{12} \mu_1$,
$L_1 = \mu_1 + w_{11}^{-1} w_{12} \mu_2$,
and 
$L_2 = \mu_2$.
By our assumptions $z_1$ and $z_2$ are independent
and $L_1$ and $L_2$ are independent.
Furthermore, rewriting $L_1$ and $L_2$ in terms of $z_1$ and $z_2$,
we get,
$L_1 = w^{-1}_{11} w^{-1}_{22} (w_{11} w_{22} - w^2_{12}) z_1 
+ w_{11}^{-1} w_{12} z_2$
and
$L_2 = z_2 -  w_{22}^{-1} w_{12} z_1$.
All linear coefficients in this transformation
are non zero due to the fact that $W$
is positive definite and that $w_{12}$ is not zero.
Consequently, due to the Skitovich-Darmois theorem,
$z_1$ is normal and $z_2$ is normal. 
Furthermore, since $z_1$ and $z_2$ are independent,
their joint pdf is normal as well.
Finally, $\{\mu_1, \mu_2\}$ and $\{z_1,z_2\}$ are
related through a non-singular linear transformation
and so $\{\mu_1,\mu_2\}$ also have a joint normal distribution
$f(\mu) = N(\mu|\eta,A)$
where $A= (a_{ij})$ is a $2 \times 2$ precision matrix.
Substituting this solution into
Equation~\ref{eq:normal1} and comparing the coefficients of 
$\mu_1^2$, $\mu_2^2$, and $\mu_1 \mu_2$, we obtain
$a_{12}/a_{11} = w_{12}/w_{11}$ and $a_{12}/a_{22} = w_{12}/w_{22}$.
Thus $A= \gamma W$ where $\gamma >0$.
\qed

The proofs of Theorems~\ref{thm:Wishart} and~\ref{thm:normal}
can be combined to form the following characterization
of the normal-Wishart distribution.

\begin{theorem} \label{thm:normalWishart}
Let $W$ be an $n \times n$, $n \ge 3$, positive-definite symmetric 
matrix of real random variables such that no entry in $W$ is zero,
$\mu$ be an $n$-dimensional vector of random variables,
and $f(\mu, W)$ be a joint pdf of $\{\mu, W\}$.
Then, $f(\mu,W)$ is 
an $n$ dimensional normal-Wishart distribution 
if and only if
$\{\mu_1, W_{11} - W_{12} W_{22}^{-1} W'_{12} \}$ 
is independent of 
$\{\mu_2 + W_{22}^{-1} W'_{12} \mu_1, W_{12},W_{22}\}$
for every partitioning $\mu_1, \mu_2$ of $\mu$ where
$W_{11}$,$W_{12}$, $W'_{12}$, $W_{22}$ is a
block partitioning of $W$ compatible
the partitioning $\mu_1, \mu_2$.
\end{theorem} 

\noindent {\bf Proof sketch:} 
The two independence conditions,
$\{\mu_1, W_{11} - W_{12} W_{22}^{-1} W'_{12} \}$ 
independent of 
$\{\mu_2 + W_{22}^{-1} W'_{12} \mu_1, $ $W_{12},W_{22}\}$
and 
$\{\mu_2, W_{22} - W'_{12} W_{11}^{-1} W_{12} \}$ 
independent of 
$\{\mu_1 + W_{11}^{-1} W_{12} \mu_2,$ $W'_{12},W_{11}\}$,
are equivalent to the following functional equation
\begin{eqnarray}
\label{eq:nw1}
\nonumber
f(\mu,W) & = &
f_1(\mu_1, \A - \B \D^{-1} \B') 
f_{2 \vb 1}(\mu_2 + \D^{-1} \B' \mu_1,\D,\B) 
\\
&  = & f_2(\mu_2,\D - \B' \A^{-1} \B) 
f_{1 \vb 2}(\mu_1 + \A^{-1} \B \mu_2,\A,\B)
\end{eqnarray}
where a subscripted $f$ denotes a pdf.
We show that the only solution for $f$ that satisfies
this functional equation is the normal-Wishart distribution.
Setting $W$ to a fixed value yields Equation~\ref{eq:normal1}
the solution of which is
\begin{eqnarray} \label{eq:nw2}
f(\mu,W) & \propto & N(\mu | \eta(W), \gamma(W) \cdot W) \\ \nonumber
  & = & N(\mu_2|\eta_2(W),\gamma(W) \cdot
             [W_{22} - W'_{12} W_{11}^{-1} W_{12}]) 
     \\ \nonumber
  & & \ \ \ \ \ \ \ \ \ 
      \cdot N(\mu_1|\eta_1(W)+\eta_2(W) \cdot 
        W^{-1}_{11}W_{12}-W_{11}^{-1} W_{12} \mu_2, \gamma(W) \cdot W_{11})
\end{eqnarray}
where both $\gamma$ and $\eta=(\eta_1,\eta_2)$ potentially can be
functions of $W$.  To see that these quantities in fact do not depend
on $W$, first note that the normal distributions for $\mu_2$ and
$\mu_1$ in Eq.~\ref{eq:nw2} must be proportional to the functions
$f_2$ and $f_{1 \vb 2}$ in Eq.~\ref{eq:nw1}, respectively.  Comparing
the form of $f_2$ with the normal distribution for $\mu_2$, we see
that $\gamma(W)$ and $\eta_2(W)$ can only depend on $W_{22} - W'_{12}
W_{11}^{-1} W_{12}$.  Comparing the form of $f_{1 \vb 2}$ with the
normal distribution for $\mu_1$, we see that $\gamma(W)$ and
$\eta_2(W)$ can only depend on $\{W_{11},W_{12}\}$.  Consequently,
$\gamma(W)$ and $\eta_2(W)$ must be constant.  Similarly, $\eta_1(W)$
must be a constant.  Substituting these solutions into
Equation~\ref{eq:nw1} and dividing by the common terms which are equal
to $f(\mu | W)$ yields Equation~\ref{eq:wish1}, the solution of which
for $f$ is a Wishart pdf.  \qed

Note that the conditions set on $W$
in Theorem~\ref{thm:normalWishart}, namely,
a positive-definite symmetric 
matrix of real random variables such that no entry in $W$ is zero,
are necessary and sufficient in order for $W$ to be 
a precision matrix of a complete Gaussian DAG model.

\section{Local versus Global Parameter Independence}

We have shown that the only pdf for $\{\mu,W\}$  
which satisfies global parameter independence, when the number
of coordinates is greater than two, is the normal-Wishart
distribution. We now discuss additional independence assertions
implied by the assumption of 
global parameter independence.

Consider the parameter prior for
$\{m_n, b_n,v_n\}$ when the prior for $\{\mu,W\}$
is a normal-Wishart as specified by 
Equations~\ref{eq:cln} and~\ref{eq:normalization-constant}.
By a change of variables, we get
\begin{eqnarray*}
\lefteqn{
f_n(m_n, b_n,v_n) = }
\\
\nonumber
& & 
\W(1/v_n \;|\; \alpha+n-1, T_{22} - T'_{12} T_{11}^{-1} T_{12}) 
\cdot 
N(b_n \;|\; T_{11}^{-1} T_{12}, T_{22}/v_n)
\cdot 
N(m_n \;|\; \nu_n, \amu/v_n)
\end{eqnarray*}
where the first block ($T_{11}$) corresponds to $X_1,\ldots,X_{n-1}$
and the second one-dimensional block ($T_{22}$)
corresponds to $X_n$.
We note that the only independence assumption
expressed by this product is that $m_n$ and $b_n$ 
are independent given $v_n$.  However, by standardizing
$m_n$ and $b_n$, namely defining, 
$m^*_n = (m_n - \nu_n )/(\amu/v_n)^{1/2}$ 
and $b^*_n =   (T_{22}/v_n)^{1/2} (b_n -  T_{11}^{-1} T_{12})$,
which is well defined because $T_{22}$ is positive definite 
and $v_n >0$,
we obtain a set of parameters $(m^*_n, b^*_n, v_n)$
which are mutually independent. 
Furthermore, this mutual independence property
holds for every local family
and for every Gaussian DAG model over $X_1,\ldots, X_n$.
We call this property
the {\em standard local independence} for Gaussian DAG models.

This observation leads to the following corollary
of our characterization theorems.

\begin{corollary}
If global parameter independence holds for every 
complete Gaussian DAG model
over $X_1,\ldots,X_n$ ($n \geq 3)$, then standard local
parameter independence also holds for every complete 
Gaussian DAG model over $X_1,\ldots,X_n$.
\end{corollary}

This corollary follows from the fact that global parameter independence 
implies that, due to Theorem~\ref{thm:normalWishart}, 
the parameter prior is a normal-Wishart, and
for this prior, we have shown that
standard local parameter independence must hold.

It is interesting to note that when $n=2$, there are distributions
that satisfy global parameter independence but do not satisfy 
standard local parameter independence. In particular, a prior
for a $2 \times 2$ positive definite matrix $W$ 
which has the form $\W(W | \alpha, T) H(w_{12})$,
where $H$ is some real function and $w_{12}$ is 
the off-diagonal element of $W$,
satisfies global parameter independence (as shown in the appendix)
but need not satisfy standard local parameter independence.
Furthermore, if standard local parameter independence is assumed, then
$H(w_{12})$ must be proportional to $e^{a w_{12}}$,
which means that, for $n=2$, the only pdf
for $W$ that satisfies global and standard 
local parameter independence
is the bivariate Wishart distribution.
In contrast, for $n>2$, global parameter
independence alone implies a Wishart prior.

\section{Discussion}
\label{sec:discuss}

The formula for the marginal likelihood
applies whenever 
Assumptions~\ref{ass:cme} through~\ref{ass:pi}
are satisfied, not only for Gaussian DAG models.
Another important special case is when all variables in $\U$
are discrete and  all local distributions are
multinomial.  This case has been treated in
Heckerman et al. (1995) and Geiger and Heckerman (1997)
\nocite{HGC95ml,GH97stat} 
under the additional assumption of
local parameter independence which was introduced by
Spiegelhalter and Lauritzen (1990)\nocite{SL90}.
Our generalized derivation herein dispenses this assumption
and unifies the derivation in the discrete case with 
the derivation needed for Gaussian DAG models.

Our characterization means that
the assumption of global parameter independence 
when combined with the definition of
$\hBs$, the assumption of complete model equivalence,
and the regularity assumption, may be too restrictive.
One common remedy for this problem is to 
use a hierarchical prior $p(\theta| \eta) p(\eta)$
with hyperparameters $\eta$.
When such a prior is used for Gaussian DAG models,
our results show that
for every value of $\eta$ for which
global parameter independence holds,
$p(\theta | \eta )$ must be a normal-Wishart distribution.
The difficulty with this approach is that the marginal likelihood
no longer has closed form and therefore approximate methods
such as MCMC are usually employed to compute the marginal
likelihood.  Also the elicitation
of hierarchical priors is often difficult.
Other alternative approaches have been discussed at the end of 
Section~\ref{sec:be}.

We conclude with a technical comment.
Equation~\ref{eq:wish1}, which encodes global parameter
independence for an unknown covariance matrix,
is an interesting example of a {\em matrix functional equation}.
The domain of each unknown function is a non-singular
matrix and the range is $R$.  
A well known functional equation
of this sort is the equation 
\begin{equation}
\label{eq-mat1}
f(XY)= f(X)f(Y)
\end{equation}
where
$X$ and $Y$ are non-singular matrices.  
The general solution
of this equation is $f(X) = |X|^{\alpha}$ or
$f(X) = |X|^{\alpha} sgn(|X|)$ (e.g., \Aczel, 1966)\nocite{Ac66}.  
When the domain of $f$ is the set of positive definite matrices,
the solution is simply $f(X) = |X|^{\alpha}$.

We note that the solution of Equation~\ref{eq-mat1}
is obtained 
for matrices over arbitrary fields.  Only algebraic
manipulations are used in its proof.
It seems reasonable to believe
and interesting to investigate, whether a solution
to Equation~\ref{eq:wish1} can be obtained via
purely algebraic manipulations.
The proof technique that we have employed, however, 
especially for the base case of the induction,
uses the fact that the matrices are over the real numbers.

\section*{Acknowledgments}

We thank Chris Meek for helping us shape the definition
of DAG models and correcting earlier versions of this
manuscript, Bo Thiesson for implementing the proposed scheme,
and Jim Kajiya for his help in regard to the characterization
theorems.  
We also thank J\'{a}nos \Aczel,
Enrique Castillo, Clark Glymour,
Antal \Jarai,
Gerard Letac, Helene Masson,
Peter Spirtes, and the reviewers, for their useful suggestions.  
The first four sections of this work with different emphasizes
(including, for example, details of the derivation of 
Equation~\ref{eq:bge}) have been reported in 
\cite{GH94uai,HG95uai}. 
A short version of this  work appeared in \cite{GH99uai}.

\bibliographystyle{apalike}

\section*{Appendix}
We now characterize the pdfs of
an unknown $2 \times 2$ precision matrix
that satisfy global parameter independence.
This result has been obtained in (Geiger and Heckerman, 1998)
\nocite{GH98pms}
under additional regularity conditions.

\begin{theorem} \label{thm:BinWishart}
Let $W$ be a $2 \times 2$ positive-definite symmetric 
matrix with random entries $w_{11},w_{12}$, and $w_{22}$
and let $f(W)$ be a pdf of $W$.
Then, $f(W) = |W|^{\beta} e^{\tr\{TW\}} H(w_{12})$ 
where $H$ is a real function
if and only if
$w_{11} - w^2_{12} / w_{22}$ is independent of $\{w_{12},w_{22}\}$
and
$w_{22} - w^2_{12} / w_{11}$ is independent of $\{w_{12},w_{11}\}$.
\end{theorem}

\noindent {\bf Proof:} 
That
$w_{11} - w^2_{12} / w_{22}$ is independent of 
$\{w_{12},w_{22}\}$
whenever $f(W)$ is a Wishart distribution 
(e.g., when $H(x) = \mbox{constant}$) is a well
known fact (Press 1971, p.\ 117-119)\nocite{Press71}. 
Consequently, this claim holds for any real function $H$.
We prove the other direction by 
solving the functional equation, which is
implied by the given independence assumptions,
\begin{equation}
\label{app:functional}
f(W) = 
f_1(w_{11}-w^2_{12} / w_{22}) f_{2 \vb 1}(w_{22},w_{12}) 
=   f_2(w_{22}-w^2_{12} / w_{11}) f_{1 \vb 2}(w_{11},w_{12}) 
\end{equation}
where a subscripted $f$ denotes a pdf.
To solve this functional equation, 
namely to find all pdfs
that satisfy it, we use techniques described in \cite{Ac66} 
and results from \cite{Ja86,Ja98}.

Let $w_{12}$ be a value such that the integral of
$f_{2 \vb 1}(x,w_{12})$ over the domain of $x$ is not identically zero.
Such a value for $w_{12}$ exists because 
$f_{2 \vb 1}(x,w_{12})$ integrates to 1 over its domain.
Without loss of generality, suppose this value
of $w_{12}$ is 1, 
lest we can modify the scale using the transformations
$w_{11} \leftarrow w_{12} w_{11}$
and
$w_{22} \leftarrow w_{12} w_{22}$.
We rewrite Equation~\ref{app:functional}
as
\begin{equation}
\label{app:functiona1}
f_1(w_{11}- 1 / w_{22})  f_{2 \vb 1}(w_{22},1) 
=   f_2(w_{22}- 1 / w_{11})  f_{1 \vb 2}(w_{11},1).
\end{equation}

We claim that all density functions satisfying
Equation~\ref{app:functiona1} must be positive everywhere
and smooth.
This is shown in Lemmas~\ref{right-to-derive} and ~\ref{lem_pe}
at the end of the proof.
Consequently,
we can take the logarithm of Equation~\ref{app:functional}
and then take derivatives.
First we take the logarithm and rename the functions. 
We get
\begin{equation}
\label{app:functiona2}
g_1(w_{11}- 1 / w_{22}) + g_{2 \vb 1}(w_{22}) 
=   g_2(w_{22}- 1 / w_{11}) + g_{1 \vb 2}(w_{11}) 
\end{equation}
where $g_1(x) = \ln f_1(x)$,
$g_{2 \vb 1}(x) =  \ln f_{2 \vb 1}(x,1)$,
and where $g_2$ and $g_{1 \vb 2}$ are defined analogously.

We take a mixed second derivative with respect to $w_{11}$ 
and $w_{22}$ of Equation~\ref{app:functiona2}.
We get
\begin{equation}
\label{app:functiona3}
g''_1(w_{11}- 1 / w_{22})
/w^2_{22}
=
g''_2(w_{22}- 1 / w_{11})
/w^2_{11}
\end{equation}
By substituting $w_{11}=w_{22}$ we obtain
$g''_1 = g''_2$. We denote this function by $h$ and so,
\begin{equation}
\label{app:functiona4}
w^2_{11}
h(w_{11}- 1 / w_{22})
=
w^2_{22}
h(w_{22}- 1 / w_{11})
\end{equation}
It is easy to show, using this functional 
equation for $h$,
that if $h$ were zero at some point then $h$ must
be identically zero, if $h$ is positive at one point 
then $h$ is positive everywhere, and
if $h$ is negative at one point 
then $h$ is negative everywhere.
We now take a derivative wrt $w_{11}$ 
and a derivative with respect to $w_{22}$
\[
2 w_{11} h(w_{11}- 1 / w_{22})
+
w_{11}^2 h'(w_{11}- 1 / w_{22})
=
\{w_{22}/w_{11}\}^2 h'(w_{22}- 1 / w_{11})
\]
\[
2 w_{22} h(w_{22}- 1 / w_{11})
+
w_{22}^2 h'(w_{22}- 1 / w_{11})
=
\{w_{11}/w_{22}\}^2 h'(w_{11}- 1 / w_{22}).
\]
From these equations, and using Equation~\ref{app:functiona4},
we get
\[
2 (w_{22} + 1 / w_{11}) h(w_{22}- 1 / w_{11})
=
- (w^2_{22}- 1 / w^2_{11}) h'(w_{22}- 1 / w_{11})
\]
Consequently,
\[
h'(x)/h(x) = -2/x
\]
where $x=w_{22}- 1/w_{11}$. This equation holds
for every $x \in R^+$.
Assuming $h$ is positive everywhere, we have $(\ln h(x))'= -2/x$
and so $\ln h(x) = \ln x^{-2} + c'$ where $c'$ is 
a constant. 
If $h$ is negative everywhere, we have $(\ln -h(x))'= -2/x$
and so $\ln(-h(x)) = \ln x^{-2} + c'$.
Consequently, whether $h$ is positive everywhere, 
negative everywhere, or identically zero,
it has the form $h(x)= c/x^2$ where $c$ is a constant.
Recall that $h = \left(\ln f_1\right)''$.
Hence, 
$f_1(x) = c_1 x^{-c} e^{c_2 x}$ and similarly
$f_2(x) = c'_1 x^{-c} e^{c'_2x}$ (i.e., one-dimensional
Wishart distributions with the same degrees of freedom).
We conclude by substituting $f_1$ and $f_2$
into Equation~\ref{app:functional} and proceeding
as in Equations~\ref{eq:wish2} through~\ref{eq:wish4}.
\qed

The next lemma shows that
every positive everywhere pdf
that satisfies Equation~\ref{app:functiona1} must be smooth.
Our lemma is an immediate consequence of
\Jarais Theorem which we now state.

\begin{theorem}[\Jarai, 1986,1998] \label{Ja98}

Let $X_i$ be an open subset of $R^{r_i}$ $(i=1,2,\ldots,n)$,
$T$ be an open subset of $R^s$, $Y$ be an open subset of $R^k$, 
$Z_i$ be an open subset of $R^{m_i}$
$(i=1,2,\ldots,n)$, $D$ be an open subset of $T \times Y$ and let
$Z$ be an Euclidean space.
Consider the functions $f: T \rightarrow Z$, 
$g_i: D \rightarrow X_i$, $f_i : X_i \rightarrow Z_i$, 
$h_i: D \times Z_i \rightarrow Z$, (i=1,2,\ldots,n).
Suppose that $0 \leq p \leq \infty$
and
\begin{itemize}
\item[(i)] for each $(t,y) \in D$,
\[
f(t) = \sum_{i=1}^n h_i(t,y,f_i(g_i(t,y)));
\]
\item[(ii)] $h_i$ is $p+1$ times 
continuously differentiable $(1 \leq i \leq n)$;
\item[(iii)] $g_i$ is $p+2$ times continuously differentiable
and for each $t \in T$ there exists a $y\in Y$ such that
$(t,y) \in D$ and $\frac{\partial{g_i}}{\partial{y}}(t,y)$ has
rank $r_i$, $1 \leq i \leq n$.
\end{itemize}
Then
\begin{itemize}
\item[(iv)] if $f_i$ $(i=1,2,\ldots,n)$ 
is Lebesgue measurable and (ii),(iii) are
satisfied with $p=0$ then $f$ is continuous on $T$;
\item[(v)] if $f_i$ $(i=1,2,\ldots,n)$ 
is continuous and (ii),(iii) are
satisfied with $p=0$ then $f$ is continuously differentiable
on $T$;
\item[(vi)] if $f_i$ $(i=1,2,\ldots,n)$ 
is $p$ times continuously differentiable and (ii),(iii) are
satisfied then $f$ is $p+1$ 
continuously differentiable on $T$.
\end{itemize}
\end{theorem}

This theorem is stated in \cite{Ja98} and its proof
is based on Theorems 3.3, 5.2, and 7.2 of \cite{Ja86}.
A simple corollary of \Jarai's theorem is the following.

\begin{lemma}
\label{right-to-derive}
Every Lebesgue measurable real functions
$l_1,l_2,l_{1 \vb 2}$ and $l_{2 \vb 1}$
defined on $R^+$
which satisfy
\begin{equation}
\label{app:functiona9}
l_1(y- 1 / t ) + l_{2 \vb 1}(t) 
=   l_2(t - 1 / y ) + l_{1 \vb 2}(y) 
\end{equation}
for every $y,t > 0$ such that $yt > 1$,
are $p$ times continuously differentiable 
where $p$ is arbitrary large.
\end{lemma}

\noindent {\bf Proof:}
The proof follows closely the lines 
of reasoning that \Jarai (1998)
applied to another functional equation.

Using statement $(iv)$ of Theorem~\ref{Ja98} we show
that $l_{2 \vb 1}$ is continuous. To match \Jarai's 
theorem notation we define $f= l_{2 \vb 1}$,
$f_1 = -l_1$, $f_2 = l_2$, $f_3= l_{1 \vb 2}$,
$h_i(t,y,w) = w$ for $i=1,2,3$,
$g_1(t,y) =  (y -1/t)$, 
$g_2(t,y) =  (t -1/y)$, and $g_3(t,y) = y$.
The only non obvious condition to check is that
for each $t \in R^+$ there exists a $y\in R^+$ such that
$ty>1$ and $\frac{\partial{g_i}}{\partial{y}}(t,y)$ has
rank $r_i$, $1 \leq i \leq n$.
But here the rank is 1 and so we just need to observe
that there exists a $y$
such that $\frac{\partial{g_i}}{\partial{y}}(t,y)$ is not
zero.

To show that $l_1$ is continuous
rewrite Equation~\ref{app:functiona9} as
\begin{equation}
\label{app:functiona10}
l_1(t) + l_{2 \vb 1}(y) 
=   l_2(\frac{ty^2}{ty+1}) + l_{1 \vb 2}(t + 1/y) 
\end{equation}
where $t,y >0$.
Now define
$f = l_1$, 
$f_1 = -l_{2 \vb 1}$,
$f_2 = l_2$, $f_3= l_{1 \vb 2}$,
$h_i(t,y,w) = w$ for $i=1,2,3$,
$g_1(t,y) = y$,
$g_2(t,y) = \frac{ty^2}{ty+1}$, and 
$g_3(t,y) = t+ 1/y$.
Observe that the conditions
of \Jarai's theorem hold and so $f = l_1$
is continuous.
By the symmetry of the equation, $l_2$ and $l_{1 \vb 2}$ are
also continuous on $R^+$.

Now we can apply statement (v) of \Jarai's Theorem.
We obtain, in the same way as above, that all four
functions are continuously differentiable.
Finally, applying statement (vi) of \Jarai's Theorem
in the same way, we get that all four functions
are 
twice continuously differentiable.
Repeating this process shows that all
four functions are 
$p$ times continuously differentiable 
for every $p >0$. \qed

The next Theorem and lemma show that every pdf
that satisfies Equation~\ref{app:functiona1} 
must be positive everywhere
and so taking the logarithm of this equation, 
as we have done, is legitimate.
We denote by $\lambda^s$ the s-dimensional Lebesgue measure
and by $\lambda$ the one dimensional Lebesgue measure.
  
\begin{theorem}[\Jarai, 1995, 1998] 
\label{Ja95}
Let $X_1, \ldots, X_n$ be orthogonal subspaces of $R^r$
of dimensions $r_1, \ldots r_n$, respectively. Suppose that
$r_i  \geq 1$ $(1 \leq i \leq n)$ and
$\sum_{i=1}^n r_i = r$.
Let $U$ be an open subset  of $R^r$ and 
$F: U \rightarrow R^m$ be a continuously differentiable 
function. For each $x \in U$, let $N_x$ denote the nullspace of
$F'(x)$. Let $p_i$ denote the orthogonal projection of $X$ onto
$X_i$. Suppose that $dim N_x = r-m$ and
$p_i(N_x)= X_i$ $(i = 1,\ldots, n)$ for all $x \in U$.
Let $A_i$ be a subset of $X_i$ $(i=1,\ldots,n)$.
If $A_1 \times A_2 \times \ldots \times A_n \subset U$,
and $A_i$ 
is $\lambda^{r_i}$ measurable with 
$\lambda^{r_i}(A_i) > 0$ $(1 \leq i \leq n)$, then 
$F(A_1 \times A_2 \times \ldots \times A_n)$
contains a non-empty open set.
\end{theorem}

Recall that if $X_1,\ldots,X_n$ are the standard orthogonal
axis of $R^n$, then $p_i(X_1,\ldots,X_n) = X_i$, and
$P_i(N_x) = \{x | (X_1,\ldots,X_{i-1},x,X_{i+1},\ldots,X_n) \in N_x\}$.

\begin{lemma}
\label{lem_pe}
Let $f,g,h,k$ be non-negative real functions
that are Lebesgue integrable with integral $c > 0$.
If these functions satisfy
\begin{equation}
\label{app:proof}
f(s- 1 / t )  g(t) =   h(t - 1 / s )  k(s) 
\end{equation}
for every $s,t > 0$ such that $st > 1$,
then they are everywhere positive.
\end{lemma}

\noindent {\bf Proof:}
The proof follows closely the lines 
of reasoning that \Jarai  (1998)
applied to another functional equation.

Let $\{f=0\}$ denote the set of points in the domain of $f$
for which $f$ is zero and let $\{f \neq 0\}$ denote the 
complementary set of all points in the domain for which
$f$ is not zero---namely, the set of points for which $f$ 
is positive.  
Similar notation is used for the functions $g,h$ and $k$.
The idea of the proof is to show that the set
$\{f=0\}$ and the set $\{f \neq 0\}$ 
are both open and therefore, since the domain 
of $f$ is connected, one of these sets must be empty.
The set $\{f \neq 0\}$ cannot be empty because
$f$ is non-negative and integrates to a positive constant
and so $\{f=0\}$ must be empty as claimed by the Theorem.
Similar arguments show that $g,h$ and $k$ are 
also positive everywhere.

The proof proceeds in three steps.
First we use Theorem~\ref{Ja95} to establish
that the set $\{g \neq 0 \}$ contains a non-empty
open set (i.e., it contains an inner point).
Then we show that every point in $\{f \neq 0 \}$
is an inner point and so $\{f \neq 0 \}$ is open.
Finally we show that every point in $\{f = 0 \}$
is an inner point and so $\{f = 0 \}$ is open as well.
Similar arguments work for $g,h$ and $k$.

We start by rewriting
Equation~\ref{app:proof}
in two symmetric ways.  First as
\begin{equation}
\label{app:proof1}
f(y)  g(z) =   h(x(y,z))  k(w(y,z)) 
\end{equation}
for all $y >0$, and $z>0$
where $x(y,z) = y z^2/(yz+1)$ and $w(y,z) = y + 1/z$.
Second as
\begin{equation}
\label{app:proof2}
f(y(x,w))  g(z(x,w)) =   h(x)  k(w) 
\end{equation}
for all $x >0$, and $w>0$
where $y(x,w) = x w^2/(xw+1)$ and $z(x,w) = x + 1/w$.

Step I.  We show that $\{g \neq 0\}$ contains
an inner point.
Since both $h$ and $k$ integrate to a positive constant,
there must exist two $\lambda$ measurable sets $A_h$ in
$\{h \neq 0\}$ and $A_k$ in $\{k \neq 0\}$ such that
$\lambda(A_h) > 0$ and $\lambda(A_k) > 0$.
The image of these sets under $z(x,w) = x + 1/w$
contains an inner point $z$ according to Theorem~\ref{Ja95}.
This theorem is applicable because the nullspace
of $z'$ is $\{ a (1/w^2, 1)) | a > 0\}$ and its projection
on either of the two coordinates is $R^+$.
Due to Equation~\ref{app:proof2},
and because the right hand side is not zero for any $x \in A_h$ 
and $w \in A_k$, each term on the left hand side is 
also not zero. Consequently, their image under $z(x,w)$,
which includes an inner point, belongs to $\{g \neq 0\}$.

Step II.  Let $y$ be an arbitrary point in $\{f \neq 0\}$.
We now show that $y$ is an inner point and so $\{f \neq 0\}$
is open. Let $z$ be an inner point in $\{g \neq 0\}$.
It follows that the image 
of a sufficiently small open set containing $z$
under $x(y,z) = y z^2/(yz+1)$ and under $w(y,z) = y + 1/z$
are open sets. These images belong to $\{h \neq 0\}$
and $\{k\neq 0\}$, respectively,
because the left hand side of Equation~\ref{app:proof1}
is positive. Now we fix $x$ in the image
and vary $w$ in a small open neighborhood.  Then $y$ is varied
in a small open neighborhood. Since the right hand side 
of Equation~\ref{app:proof1}
is positive, the neighborhood of $y$ belongs to
$\{f \neq 0\}$ and so $y$ is an inner point.
Similar arguments show that $\{g \neq 0\}$ is open as well.
By the symmetry of Equation~\ref{app:proof} 
the same claim holds for $h$ and $k$.

Step III.
Let $y$ be an arbitrary point in $\{f = 0\}$.
We now show that $y$ is an inner point and so $\{f = 0\}$
is open. Let $z$ be an inner point in $\{g \neq 0\}$.
It follows that the image 
of a sufficiently small open set containing $z$
under $x(y,z) = y z^2/(yz+1)$ and under $w(y,z) = y + 1/z$
are open sets. 
Since the left hand side of Equation~\ref{app:proof1}
is zero, at least one term in the right hand side
must be zero.  If $x$ is in $\{h = 0\}$, then
fix $x$. As we vary $w$ in a small open neighborhood in the image,
$g$ remains positive due to continuity. 
Also $y$ is varied in a small open neighborhood. 
Since the right hand side 
of Equation~\ref{app:proof1}
is zero, the neighborhood of $y$ belongs to
$\{f = 0\}$ and so $y$ is an inner point.
The other case is when $w$ is in $\{k = 0\}$, in
which case we fix $w$ and vary $x$ in a small neighborhood.
Similar arguments show that $\{g = 0\}$ is open as well.
By the symmetry of Equation~\ref{app:proof},
the same claim holds for $h$ and $k$.
\qed

Note that Theorems~\ref{right-to-derive} and~\ref{lem_pe}
together imply that every pdf that solves 
Equation~\ref{app:proof} must be positive everywhere and
smooth.  

\end{document}